\documentclass[conference]{IEEEtran}
\IEEEoverridecommandlockouts
\usepackage{amsmath,amssymb,amsfonts}
\usepackage{algorithmic}
\usepackage{graphicx}
\usepackage{textcomp}
\usepackage{xcolor}
\usepackage{subcaption}
\usepackage[T1]{fontenc}
\usepackage{booktabs}
\usepackage{marvosym}
\usepackage[marginal]{footmisc}
\usepackage{hyperref}
\usepackage{float}
\usepackage{cite}
\usepackage{hyperref}
\usepackage[inkscapelatex=false]{svg}
\hypersetup{
colorlinks=true,
linkcolor=black,
citecolor=black,
urlcolor=black
}
\usepackage{arydshln}
\usepackage{dashrule}

\def\BibTeX{{\rm B\kern-.05em{\sc i\kern-.025em b}\kern-.08em
    T\kern-.1667em\lower.7ex\hbox{E}\kern-.125emX}}

    \makeatletter
\newcommand{\linebreakand}{
  \end{@IEEEauthorhalign}
  \hfill\mbox{}\par
  \mbox{}\hfill\begin{@IEEEauthorhalign}
}
\makeatother
\begin{document}

\title{JurisCTC: Enhancing Legal Judgment Prediction via Cross-Domain Transfer and Contrastive Learning
}

\author{\IEEEauthorblockN{Zhaolu Kang\IEEEauthorrefmark{1}}
\IEEEauthorblockA{\textit{School of Software \& Microelectronics} \\
\textit{Peking University}\\
Beijing, China \\
kangzl9966@gmail.com}
\and
\IEEEauthorblockN{Hongtian Cai\IEEEauthorrefmark{1}}
\IEEEauthorblockA{\textit{School of Electrical \& Electronic Engineering} \\
\textit{Nanyang Technological University}\\
Singapore \\
cht7janus@gmail.com}
\and
\IEEEauthorblockN{Xiangyang Ji}
\IEEEauthorblockA{\textit{College of Software} \\
\textit{Jilin University}\\
Changchun, Jilin, China \\
jixy5523@mails.jlu.edu.cn}
\linebreakand 
\IEEEauthorblockN{Jinzhe Li}
\IEEEauthorblockA{\textit{College of Computer Science and Technology} \\
\textit{Jilin University}\\
Changchun, Jilin, China \\
lijz2128563286@outlook.com}
\and 
\IEEEauthorblockN{Nanfei Gu\IEEEauthorrefmark{2}}
\IEEEauthorblockA{\textit{KoGuan School of Law} \\
\textit{Shanghai Jiao Tong University}\\
Shanghai, China \\
gunanfei@126.com}
\thanks{\IEEEauthorrefmark{1} These authors contribute equally to this work. They are co-first authors.}
\thanks{\IEEEauthorrefmark{2} Corresponding author.}
\thanks{$^{1}$ JurisCTC is available at: \href{https://github.com/Zhaolu-K/JurisCTC}{https://github.com/Zhaolu-K/JurisCTC}}
}
\maketitle

\begin{abstract}
In recent years, Unsupervised Domain Adaptation (UDA) has gained significant attention in the field of Natural Language Processing (NLP) owing to its ability to enhance model generalization across diverse domains.
However, its application for knowledge transfer between distinct legal domains remains largely unexplored.
To address the challenges posed by lengthy and complex legal texts and the limited availability of large-scale annotated datasets, we propose JurisCTC, a novel model designed to improve the accuracy of Legal Judgment Prediction (LJP) tasks.
Unlike existing approaches, JurisCTC facilitates effective knowledge transfer across various legal domains and employs contrastive learning to distinguish samples from different domains.
Specifically, for the LJP task, we enable knowledge transfer between civil and criminal law domains.
Compared to other models and specific large language models (LLMs), JurisCTC demonstrates notable advancements, achieving peak accuracies of 76.59\% and 78.83\%, respectively.\footnotemark[1]
\end{abstract}


\section{Introduction}

Legal Judgment Prediction (LJP) refers to the task of forecasting court outcomes based on the facts of a legal case, as well as other relevant information such as arguments and claims presented in the case description.
This field aims to leverage computational techniques to predict judicial decisions, offering significant benefits across various legal contexts.
Automated LJP systems have considerable practical value: they can assist legal professionals in analyzing cases and providing consultation services to the public, thereby reducing legal costs and improving access to justice.

Despite the potential benefits, recent attempts have primarily focused on the text analysis of judgments and the prediction of specific legal domains.
These approaches often neglect the relationship between the judicial outcomes and the logical consistency of the legal applications.
Most of these works extract efficient features from text (e.g., N-grams) or case profiles (e.g., dates, terms, locations, and types).
These methods require substantial human effort for feature engineering and case annotation, which can be both time-consuming and resource-intensive.
Furthermore, they often face generalization challenges when applied to diverse legal scenarios, limiting their applicability across legal contexts.
This highlights the need for more robust and adaptable models to handle diverse legal data with minimal manual intervention.

Within the complex domain of legal applications, criminal law, and civil law have emerged as the most extensively explored fields of LJP.
However, they are not interoperable with each other.
In criminal law, LJP models are designed to predict outcomes, such as applicable legal articles, charges, and prison terms~\cite{Aletras2016,feng2022legal, malik2021ildc, xu2020distinguish, 2021neurjudge, 2022CEMC}.
In civil law, LJP models focus on determining whether the plaintiff’s claims can be upheld~\cite{2019Auto-Judge, 2021LJP-MSJudge}.

The effectiveness of LJP models heavily depends on the quantity and quality of judgment documents used for training.
However, due to considerations of national security and social stability, the number of criminal judgments publicly available on China Judgments Online has significantly decreased, and the shortage of data directly restricts the development of LJP research in the field of criminal law.
This data scarcity poses a significant challenge for researchers aiming to develop robust LJP models.

To address this issue, we propose leveraging transfer learning techniques.
The integration of transfer learning in LJP not only mitigates the issue of limited training data but also enhances the robustness and adaptability of predictive models.
By leveraging the rich data available in civil law, we can create more generalized models that perform well even in the constrained environment of criminal law.
Transfer learning enables the extraction of knowledge from civil law judgments, tapping into the collective expertise of judges to improve criminal law predictions.

Additionally, we incorporate unsupervised domain adaptation (UDA) into criminal law outcome prediction, exploring interoperability between different legal fields.
This approach demonstrates the potential of cross-domain learning, where insights from one legal domain can inform and enhance another.
This approach also opens up new avenues for interdisciplinary research, where insights from one legal domain can inform and improve practices in another.
Given the adjustment of the Chinese policy of disclosing judicial documents, this research method has more substantial practical value and theoretical significance.

Our contributions are summarized as follows:
\begin{itemize}
\item[$\bullet$] We propose a method to use transfer learning methods in different legal fields to improve the problem of insufficient training data, providing new ideas for research on cross-departmental laws and paving the way for subsequent knowledge transfer in other legal domains.
\item[$\bullet$] We introduce a method of cross-domain transfer and contrastive learning to improve the accuracy of LJP.
\end{itemize}

\section{Related Work}

\subsection{Legal Judgment Prediction}
In recent years, Legal Judgment Prediction (LJP) has garnered significant attention and achieved substantial progress.
The availability of extensive legal judgment data has spurred a growing body of research dedicated to this topic.
Recent advancements in Natural Language Processing (NLP) have significantly contributed to the development of LJP models.
These models leverage large-scale public datasets and sophisticated algorithms to predict judicial outcomes with impressive accuracy.
However, the complexity of legal reasoning and the subjective nature of legal arguments present ongoing challenges.
To address these issues, researchers have begun incorporating argument analysis into LJP models, enhancing their ability to evaluate the quality of legal arguments presented by the parties involved.

Current research in LJP primarily focuses on predicting case outcomes~\cite{2018RACP,2018Criminal,2020QAjudge,2022survey}, such as applicable legal articles, charges, and prison terms, based on factual information~\cite{2018DPAM,cail2018,2018TOPJUDGE}, plaintiff narratives~\cite{2019Auto-Judge,2021LJP-MSJudge}, and other relevant court-presented information~\cite{2019CPTP,2019MAMD}.

Despite the comprehensive nature of these tasks, the interdependence of prediction results often leads to a lack of intuitive clarity.
This issue is particularly pronounced in criminal law, where the complex interplay between various legal outcomes can obscure the direct verdict of guilt or innocence.
Conversely, predictions in civil law tend to be more straightforward, providing a clearer depiction of outcomes~\cite{2018DPAM,2019Auto-Judge,2021LJP-MSJudge}.
This distinction is crucial for our research, which focuses on the logical application of Chinese law.
Given the gradual decline in the availability of criminal law data, we identify a valuable opportunity to utilize civil law datasets to address this gap.

Our research aims to build on these developments by focusing on the application of LJP in the context of Chinese civil law, leveraging the unique characteristics of civil law datasets to enhance predictive accuracy and clarity.

\subsection{Unsupervised Domain Adaptation}
Deep feed-forward architectures have brought impressive advances to the state-of-the-art across a wide variety of machine learning tasks and applications.
However, these leaps in performance are typically contingent upon the availability of large amounts of labeled training data.
For problems where labeled data is scarce, it is often possible to obtain sufficiently large training sets, but these may suffer from a shift in data distribution compared to the actual data encountered at test time.
A notable example is synthetic or semi-synthetic training data, which can be abundant and fully labeled, yet inevitably differ in distribution from real-world data.

Machine learning has been widely applied in various fields~\cite{zhang2024dunet,zhang2020machine,duan2023orchestrating,zhang2024amter,vaswani2017attention,liu2022integration,zhang2021review,li2024gansamples}.
Unsupervised Domain Adaptation (UDA) has proven to be an effective strategy for transferring knowledge from a well-labeled source domain to an unseen, diverse, and unlabeled target domain.
By leveraging data from both labeled source domains and unlabeled target domains, UDA facilitates the performance of various tasks within the target domain.
This approach has been successfully applied in several areas, including natural image processing, video analysis, natural language processing, time-series data analysis, and medical image analysis.

In the realm of NLP, the development of UDA methods has become increasingly crucial, particularly due to the prohibitive costs associated with annotating extensive language datasets.
UDA techniques have been utilized for a spectrum of NLP tasks, such as sentiment analysis~\cite{ramponi-plank-2020-neural, SIP-2022-0019}, relation extraction~\cite{10.1093/bioinformatics/bty190,shi-etal-2018-genre}, and language identification~\cite{li-etal-2018-whats}.
Pioneering efforts in NLP UDA include the Domain-Adversarial Neural Networks (DANN) proposed by Ganin et al.~\cite{DANN}, which achieve UDA by integrating a domain classifier with the feature extractor via a gradient reversal layer. 

The primary focus of UDA research is on learning domain-invariant features.
This can be achieved either by explicitly reducing the distance between source and target feature spaces using some distribution discrepancy metric or by adversarial training, where the feature extractor is trained to fool a domain classifier.
Both approaches are jointly optimized to achieve an aligned feature space.
Our research focuses on applying the latter approach in transformer-based models, such as BERT~\cite{bert2019}, for textual tasks.

\section{Methods}
\subsection{Overview}
\begin{figure}[!t]
    \centering \includegraphics[width=1\linewidth]{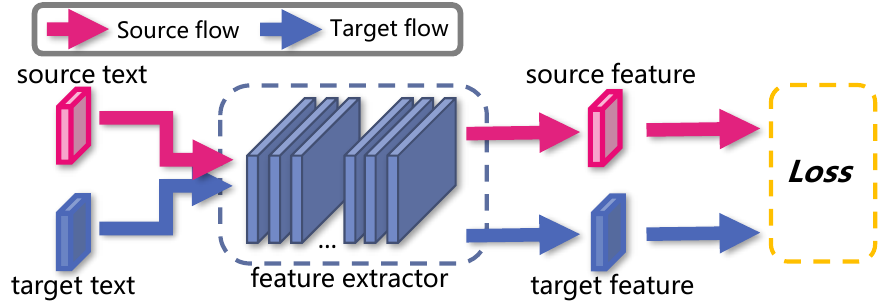}
    \caption{This diagram depicts an advanced domain adaptation model that integrates BERT for feature extraction. The architecture processes source and target data in tandem, utilizing BERT to derive features that inform loss calculations for both label prediction and domain classification.}
    \label{fig:main}
    \begin{subfigure}{0.98\linewidth}  
        \centering \includegraphics[width=1\linewidth]{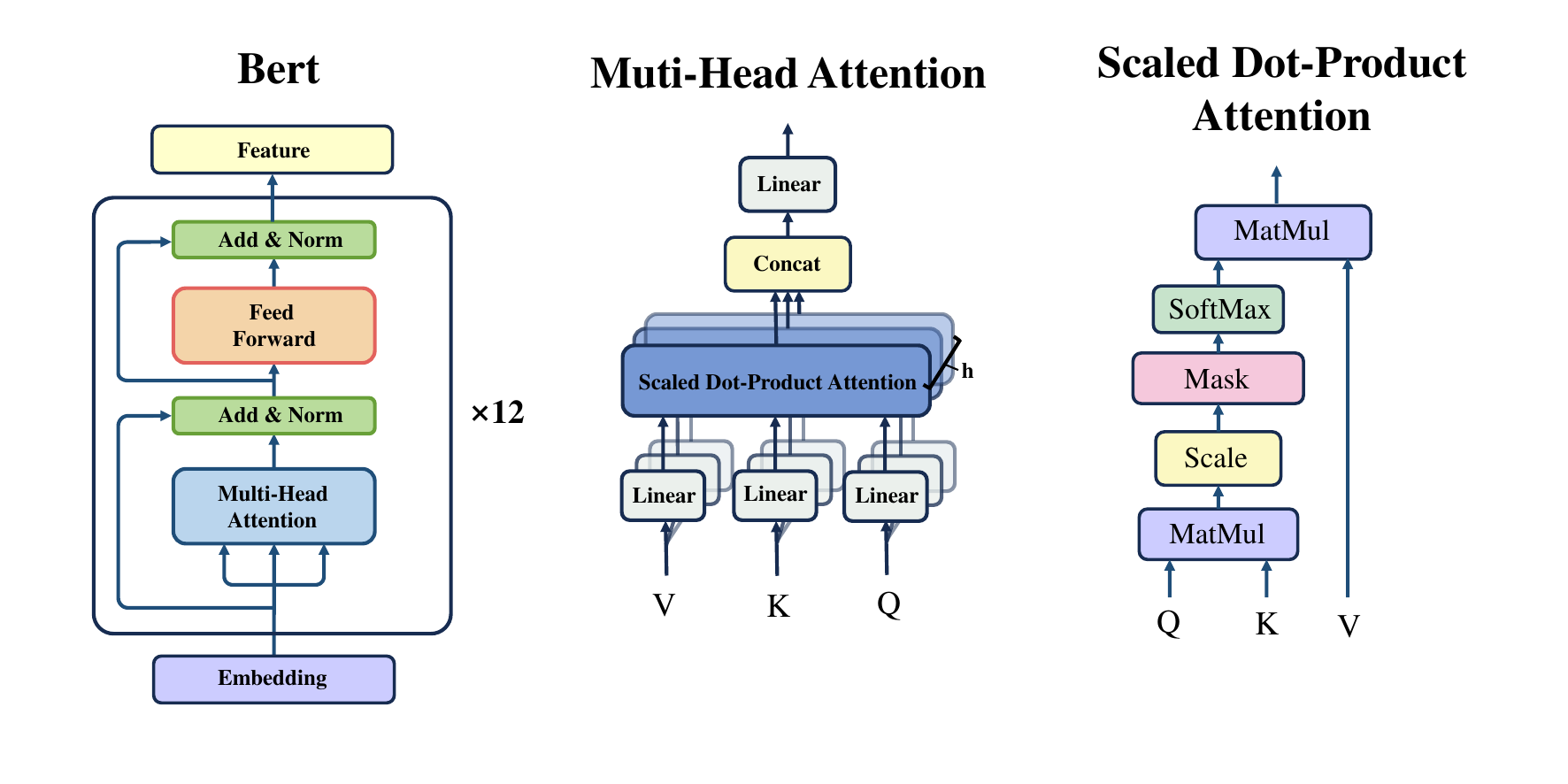}     
        \caption{Subfigure 1: The diagram delineates the intricate structure of the model, showcasing the 'Bert' block as a multifaceted feature extractor with repeated 'Add \& Norm' and 'Multi-Head Attention' processes. The 'Multi-Head Attention\' block is detailed with 'Concat' and 'Linear' operations leading into the 'Scaled Dot-Product Attention' mechanism.}    
        \label{figure:sub}
    \end{subfigure}
    
    \begin{subfigure}{0.98\linewidth} 
        \centering \includegraphics[width=1\linewidth]{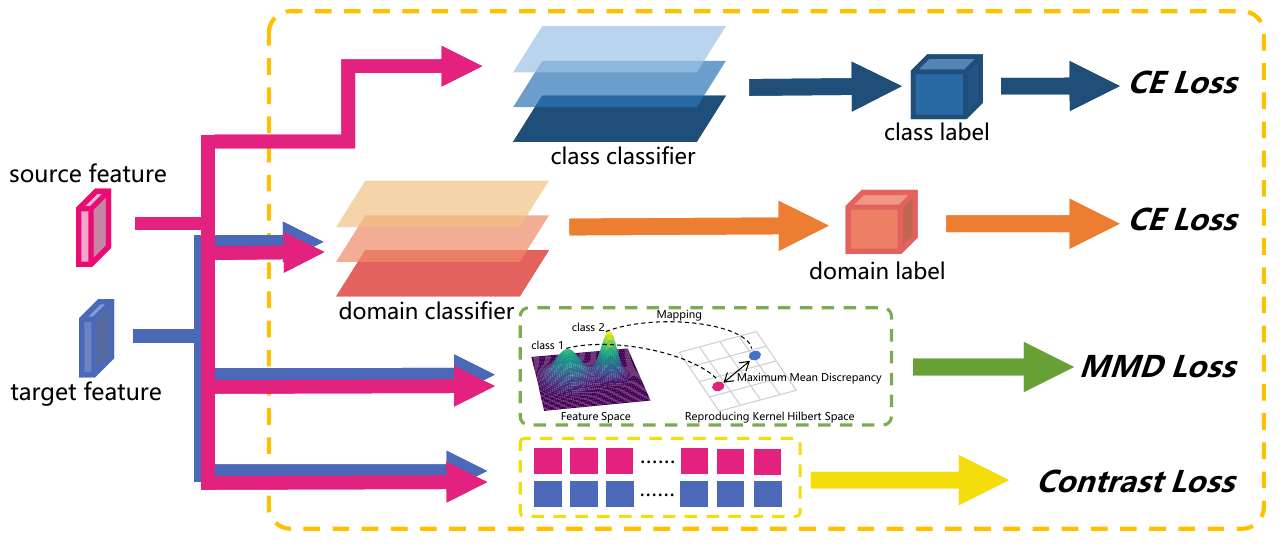}
        \caption{Subfigure 2: The diagram illustrates the loss components in a domain adaptation model. Cross-entropy (CE) Loss is used for class and domain classification. Maximum Mean Discrepancy (MMD) Loss aligns source and target feature distributions. Contrastive Loss pulls similar instances together and pushes dissimilar ones apart to enhance feature learning.}  
        \label{figure:sub2}
    \end{subfigure}
\end{figure}

Our model comprises three key components: a BERT feature extractor, a class classifier, and a domain classifier.
Figure~\ref{fig:main} illustrates the overall architecture of our model, highlighting the interactions between the BERT feature extractor, the class classifier, and the domain classifier.

Initially, the BERT feature extractor interprets the input text, transforming it into a rich set of features that capture contextual and semantic intricacies.
This phase leverages the powerful pre-trained BERT model to generate embeddings that encapsulate the nuanced meanings of the input text.

In the second phase, the class classifier, a sophisticated neural network, is applied to the source domain.
This fully connected network maps the extracted features to the corresponding labels within the source domain, fine-tuning the pre-trained BERT representations to our specific classification requirements.
This step ensures that the model is well adapted to the specific task at hand, improving its accuracy in predicting the correct labels.

The third phase involves the strategic application of the domain classifier across both the source and target domains.
The domain classifier is trained to distinguish between features from the source and target domains.
By iteratively applying the domain classifier, our model progressively learns to reduce domain discrepancies through adversarial training.
This process involves a gradient reversal layer that encourages the feature extractor to produce domain-invariant features, thereby improving the predictive accuracy and robustness in domain adaptation scenarios.

\subsection{Forward Pass}

BERT's architecture is ingeniously designed to capture the intricate nuances of language by leveraging the power of bidirectional Transformer encoders.
The process begins with tokenizing input sentences into discrete tokens, which are then embedded into vectors.
These vectors are processed through multiple layers of the Transformer encoder, each refining the token representations through a series of operations.

Each encoder layer in BERT performs a sequence of operations on the input embeddings.
The first operation within each layer is the multi-head self-attention mechanism, which allows the model to consider each word in the context of the entire sentence.
This is achieved by generating query(Q), key(K), and value(V) vectors for each token and computing attention scores that determine the influence of other tokens.

The multi-head attention is computed as follows:
\begin{equation}
\text{MultiHead}(Q, K, V) = \text{Concat}(\text{head}_1, \ldots, \text{head}_h)W^O
\end{equation}

Where each head \( \text{head}_i \) is computed as:

\begin{equation}
\text{head}_i = \text{Attention}(QW_i^Q, KW_i^K, VW_i^V)
\end{equation}

The mathematical formulation of the self-attention mechanism is as follows:
\begin{equation}
\text{Attention}(Q, K, V) = \text{softmax}\left(\frac{QK^T}{\sqrt{d_k}}\right)V
\end{equation}

Where \( d_k \) is the dimensionality of the keys.
This is done multiple times in parallel for each 'head' in the multi-head attention, allowing the model to capture different aspects of the data.

After the attention mechanism, each layer independently applies a position-wise feed-forward network to each tokey.
This network comprises two linear transformations with a ReLU activation in between.

The position-wise feed-forward network can be mathematically expressed as:
\begin{equation}
\text{FFN}(x) = \max(0, xW_1 + b_1)W_2 + b_2
\end{equation}

Where \( W_1 \), \( W_2 \), \( b_1 \), and \( b_2 \) are the parameters of the linear transformations.

During pre-training, BERT uses the MLM objective, which is designed to predict the original vocabulary ID of the masked words based on the context provided by the other non-masked words in the sequence.

The loss function for MLM is the cross-entropy loss over the vocabulary:
\begin{equation}
L_{\text{MLM}} = -\sum_{i=1}^{N} \log p(w_i | w_{\text{context}})
\end{equation}

Where \( N \) is the number of masked tokens, \( w_i \) is the true token, and \( w_{\text{context}} \) represents the surrounding non-masked tokens.

The BERT feature extractor, which serves as the cornerstone of our approach, diligently processes the input text to produce a feature vector \( f \) that encapsulates the linguistic context and semantic richness.
This vector, existing within the high-dimensional space \( \mathbb{R}^{m \times d} \), where \( m \) and \( d \) denote the maximum input sequence length and the hidden state dimension respectively, forms the substrate for the subsequent predictive tasks.

By leveraging the pre-trained BERT model and fine-tuning it on our domain-specific corpus, we adapt its extensive linguistic knowledge to our particular use case.
The model’s proficiency in discerning contextual dependencies is not merely confined to adjacent tokens but extends to encompass long-range dependencies, thereby mitigating the limitations traditionally associated with sequential processing models.
This enables the extraction of features that are highly predictive of the outcomes of interest, thus enhancing the performance of our predictive models.

The class classifier, structured as a neural network with multiple layers, then takes over.
It processes the feature vector \( f \), applying a series of transformations that culminate in the prediction of the appropriate labels.
This is achieved through a function \( g: \mathbb{R}^{m \times d} \rightarrow \mathbb{R}^l \), where \( l \) is the number of possible labels. 

The function \( g \) is defined as follows:
\begin{equation}
g(f) = \sigma(W_l f + b_l)
\end{equation}

Where \( W_l \) is the weight matrix, \( b_l \) is the bias vector, and \( \sigma \) is the activation function that introduces non-linearity, enabling the network to learn complex patterns.

Simultaneously, the domain classifier, another neural network, engages in a parallel process.
It assesses the feature vector \( f \) to determine the domain of origin, employing a function \( h: \mathbb{R}^{m \times d} \rightarrow \mathbb{R}^k \), with \( k \) representing the number of domains.

The function \( h \) is expressed as:
\begin{equation}
h(f) = \sigma(W_d f + b_d)
\end{equation}

In this equation, \( W_d \) is the domain-specific weight matrix, and \( b_d \) is the corresponding bias vector.
The activation function \( \sigma \) again plays a vital role in facilitating the ability to differentiate between domains.

\subsection{Error Backpropagation}

To achieve domain adversarial training, we implement a gradient reversal layer (GRL).
The GRL is a unique component that facilitates the alignment of feature distributions between the source and target domains.
It operates without any trainable parameters, except for a meta-parameter $\lambda$, which is not updated by backpropagation.
During the forward pass, the GRL acts as an identity function, allowing the data to pass through unchanged.
However, during backpropagation, the GRL multiplies the gradient by $\lambda$ and reverses its direction, effectively encouraging the feature extractor to produce domain-invariant features.

Mathematically, the GRL can be represented as:
\begin{equation} 
\text{GRL}(x) = x \quad 
\text{(forward pass)} 
\end{equation}
\begin{equation} 
\frac{\partial 
\text{GRL}(x)}{\partial x} = -\lambda \quad \text{(backward pass)} 
\end{equation}

The loss function for the domain classifier, \( L_d \), is defined using binary cross-entropy loss.
This loss measures the discrepancy between the predicted domain labels and the true domain labels, guiding the model to distinguish between the source and target domains accurately.

The binary cross-entropy loss is given by:
\begin{equation} L_d = -\frac{1}{N} \sum_{i=1}^{N} [y_i \log(p_i) + (1 - y_i) \log(1 - p_i)] \end{equation}

Where N is the number of samples, \( y_i \) is the true domain label, and \( p_i \) is the predicted probability of the sample belonging to the source domain.

The classification loss, \( L_{y} \), is computed on the source domain using cross-entropy loss between the predicted labels and the true labels, ensuring the predictive accuracy on the source domain tasks. 

The cross-entropy loss for classification is defined as:

\begin{equation} L_y = -\frac{1}{N} \sum_{i=1}^{N} \sum_{c=1}^{C} y_{i,c} \log(p_{i,c}) \end{equation}

Where \( C \) is the number of classes, \( y_{i,c} \) is a binary indicator (0 or 1) if class label \( c \) is the correct classification for sample \( i \), and \( p_{i,c} \) is the predicted probability of sample \( i \) being in class \( c \).

To further align the feature distributions of the source and target domains, we employ the Maximum Mean Discrepancy (MMD) method.
MMD is a non-parametric measure that quantifies the distance between the feature distributions of the two domains. By incorporating MMD loss into our training process, we encourage the model to minimize domain discrepancies.

The MMD is based on the Gaussian Kernel, defined as:

\begin{equation}
k(x, y) = \exp\left(-\frac{\|x - y\|^2}{2\sigma^2}\right)
\end{equation}

where \(k(x, y)\) is the Gaussian Kernel between two samples \(x\) and \(y\), \(\|x - y\|^2\) is the squared Euclidean distance between the samples, and \(\sigma\) is the bandwidth parameter controlling the kernel's spread.

The MMD loss is formulated as:

\begin{equation}
L_{\text{MMD}} = \left\| \frac{1}{n_s} \sum_{i=1}^{n_s} \phi(\mathbf{f}_i^s) - \frac{1}{n_t} \sum_{j=1}^{n_t} \phi(\mathbf{f}_j^t) \right\|^2
\end{equation}

Where $\phi$ denotes a feature map projecting features into a reproducing kernel Hilbert space, $\mathbf{f}_i^s$ and $\mathbf{f}_j^t$ are the feature representations of the source and target domains, respectively, and $n_s$ and $n_t$ are the number of samples in each domain.

Additionally, we incorporate Contrastive Learning to enhance the alignment of feature distributions between the two domains.
This method minimizes the distance between similar samples while maximizing the distance between dissimilar ones.
We leverage both source and target domain samples by extracting their features and computing a similarity matrix, capturing the relationships between all pairs of samples from the two domains.

The Contrastive Learning function can be defined as follows.

Concatenate source and target domain features:

\begin{equation}
\mathbf{z} = [\mathbf{z}_s; \mathbf{z}_t]
\end{equation}

where \(\mathbf{z}_s\) represents source domain features and \(\mathbf{z}_t\) represents target domain features.

Normalize the features:

\begin{equation}
\mathbf{z}_{\text{norm}} = \frac{\mathbf{z}}{\|\mathbf{z}\|}
\end{equation}

where \(\mathbf{z}_{\text{norm}}\) is the normalized feature vector.

Compute the similarity matrix:

\begin{equation}
\mathbf{S} = \mathbf{z}_{\text{norm}} \mathbf{z}_{\text{norm}}^\top
\end{equation}

where \(\mathbf{S}_{i,j}\) represents the similarity between sample \(i\) and sample \(j\).

Construct the contrastive learning label matrix:

\begin{equation}
\mathbf{M}_{i,j} = 
\begin{cases} 
1, & \text{if } y_i = y_j \\
0, & \text{otherwise}
\end{cases}
\end{equation}

where \(y_i\) and \(y_j\) are the labels of the \(i\)-th and \(j\)-th samples, respectively.

The Contrastive Learning Loss is defined as:

\begin{equation}
L_{\text{contrast}} = -\frac{1}{N} \sum_{i=1}^{N} \log \frac{\exp \left( \frac{\text{sim}(z_i, z_j)}{\tau} \right)}{\sum_{k=1}^{N} \exp \left( \frac{\text{sim}(z_i, z_k)}{\tau} \right)}
\end{equation}

where \( N \) is the batch size. \( z_i \) and \( z_j \) represent feature vectors of positive sample pairs, and \( \text{sim}(z_i, z_j) \) is the similarity between feature vectors \( z_i \) and \( z_j \), often computed as cosine similarity.\( \tau \) is a temperature parameter.

The total loss ($L_{\text{total}}$) is formulated as a weighted sum of the individual losses:

\begin{equation}
L_{\text{total}} = L_{y} + \lambda_d L_{d} + \lambda_{\text{MMD}} L_{\text{MMD}} + L_{\text{contrast}}
\end{equation}

Where $\lambda_d$ and $\lambda_{\text{MMD}}$ are hyperparameters that control the trade-off between the domain classification accuracy and the domain invariance of the features.

By integrating the MMD loss and Contrastive Learning into the overall loss function, we provide a robust statistical basis for the model to reduce domain variance, thus enhancing its adaptability and performance on the target domain.

\section{Experiments}

\subsection{Datasets}

In our experiments, we constructed a comprehensive dataset by integrating data from two well-known legal datasets: the LJP-MSJudge~\cite{2021LJP-MSJudge} Civil Law dataset and the CAIL-2018~\cite{cail2018} Criminal Law dataset.
This integrated dataset serves as the foundation for our experimental analysis, encompassing a diverse array of judgments from both civil and criminal law domains.
The LJP-MSJudge dataset includes a detailed collection of civil cases, each containing the plaintiff’s claims, court debate records, and judgment verdicts.
The CAIL-2018 dataset comprises fact descriptions, applicable law articles, charges, and terms of penalties for criminal cases.

Civil cases in China often undergo multiple trials, such as first instance, second instance, and retrial.
However, the findings of fact in the first instance judgment are crucial in determining the case outcome.
Given the relatively low correction rate of civil cases in China, we primarily refer to the case facts and judgment results from the first-instance judgments in the LJP-MSJudge dataset.

Our primary task focuses on predicting the outcomes of judgments.
Therefore, we categorize the text related to judgment outcomes into two distinct categories: for civil cases, the outcomes are either supporting or not supporting the plaintiff's appeal; for criminal cases, the outcomes are either guilty or not guilty. The detailed statistics of the datasets are presented in Table~\ref{tab:datasets}.

\begin{table}[!t]
\caption{Dataset Statistical Overview}
\begin{center}
\begin{tabular}{|l|c|c|c|c|}
\hline
\textbf{Dataset} & \multicolumn{2}{|c|}{\textbf{Criminal Law}} & \multicolumn{2}{|c|}{\textbf{Civil Law}} \\
\cline{2-5}
& \textbf{Guilty} & \textbf{Not Guilty} & \textbf{Support} & \textbf{Not Support} \\
\hline
\textbf{Number} & 2509 & 2227 & 2133 & 1922 \\
\hline
\end{tabular}
\end{center}
\label{tab:datasets}
\end{table}
\subsection{Experimental Settings}
For training, we set the learning rate of Adam optimizer to $10^{-3}$, and the batch size to 128.
After training every model for 16 epochs, we choose the best model on the validation set for testing.

To compare the performance of the baselines and our methods, we choose four metrics that are widely used for multi-classification tasks, including accuracy (Acc.), macro-precision (MP), macro-recall (MR), and macro-F1 (F1).

\subsection{Baselines}
To extensively validate the effectiveness of the proposed model, the following baselines are employed for comparison.
\begin{itemize}
\item[$\bullet$]\textbf{TextCNN}~\cite{textCNN} is a convolutional neural network trained on top of pre-trained word vectors for sentence-level classification tasks.
\item[$\bullet$]\textbf{BERT}~\cite{bert2019} is a fine-tuning representation model which has been applied to learn a good representation of the input fact
summary for judgment prediction. 
\item[$\bullet$]\textbf{TOPJUDGE}~\cite{2018TOPJUDGE}  is a topological multi-task learning framework for LJP, which formalizes the explicit dependencies over subtasks in a directed acyclic graph.
\item[$\bullet$]\textbf{MPBFN}~\cite{mpbfn2019} is a multi-task learning framework for LJP with multiperspective forward prediction and backward verification.
\end{itemize}

We then compare several high-performing LLM models. These models are currently among the most popular and exhibit the strongest capabilities in understanding and reasoning.
\begin{itemize}
\item[$\bullet$]\textbf{GPT-4o} is a state-of-the-art language model designed for various natural language processing tasks.
\item[$\bullet$]\textbf{Gemini-1.5-Flash} is a high-performance language model tailored for fast and accurate text generation and understanding.
\item[$\bullet$]\textbf{DeepSeek-V3-Chat} is an advanced conversational AI model optimized for dialogue and information retrieval tasks.
\end{itemize}

These baselines provide a comprehensive foundation for evaluating the proposed model’s performance.
Each baseline represents a different approach to natural language processing tasks, offering a diverse set of methodologies for comparison.

\subsection{Overall Performance}
To evaluate the performance of the proposed model, we export the results from the following perspectives:

\subsubsection{Comparison against baselines}
We conducted experiments to evaluate the performance of transfer learning from Criminal Law to Civil Law and vice versa.
Table Table~\ref{tab:results_civil_to_criminal} and Table Table~\ref{tab:results_criminal_to_civil} show the experimental results. 

\begin{table}[!t]
\caption{Performance comparison on Criminal Law tasks}
\begin{center}
\begin{tabular}{|l|c|c|c|c|}
\hline
\textbf{Model} & \textbf{Acc.} & \textbf{MP} & \textbf{MR} & \textbf{F1} \\
\hline
GPT-4o (Sep. 2024) & 66.46 & \textbf{83.00} & 51.98 & 63.93 \\
DeepSeek-V3-Chat & 67.23 & 77.75 & 53.29 & 63.24 \\
Gemini-1.5-Flash & 68.77 & 78.87 & 55.59 & 65.22 \\
\hline
BERT & 43.51 & 50.00 & 6.30 & 11.14 \\
TextCNN & 43.47 & 41.74 & 49.92 & 45.47 \\
TOPJUDGE & 44.80 & 72.04 & 51.14 & 59.82 \\
MPBFN & 43.51 & 46.75 & 49.96 & 48.30 \\
\hline
\textbf{JurisCTC} & \textbf{76.59} & 75.92 & \textbf{85.75} & \textbf{80.54} \\
\hline
\end{tabular}
\label{tab:results_civil_to_criminal}
\end{center}
\end{table}

\begin{table}[!t]
\caption{Performance comparison on Civil Law tasks}
\begin{center}
\begin{tabular}{|l|c|c|c|c|}
\hline
\textbf{Model} & \textbf{Acc.} & \textbf{MP} & \textbf{MR} & \textbf{F1} \\
\hline
GPT-4o (Sep. 2024) & 52.04 & 51.80 & 51.83 & 51.81 \\
DeepSeek-V3-Chat & 46.65 & 49.09 & 49.40 & 49.25 \\
Gemini-1.5-Flash & 59.65 & 58.56 & 55.24 & 56.85 \\
\hline
BERT & 64.37 & 63.77 & 86.97 & 73.58 \\
TextCNN & 57.74 & 56.10 & 55.59 & 55.84 \\
TOPJUDGE & 67.31 & 75.07 & 62.51 & 68.22 \\
MPBFN & 62.89 & 64.33 & 58.49 & 61.27 \\
\hline
\textbf{JurisCTC} & \textbf{78.83} & \textbf{76.59} & \textbf{90.61} & \textbf{83.01} \\
\hline
\end{tabular}
\label{tab:results_criminal_to_civil}
\end{center}
\end{table}

The experimental results demonstrate that the proposed model, JurisCTC, outperforms all baseline models in both Criminal Law and Civil Law tasks.
Specifically, JurisCTC achieves the highest accuracy, macro-recall, and macro-F1 scores across both domains.

In the Criminal Law tasks, JurisCTC achieves an accuracy of 76.59\%, significantly higher than the best-performing baseline, Gemini-1.5-Flash, which has an accuracy of 68.77\%.
Similarly, in the Civil Law tasks, JurisCTC achieves an accuracy of 78.83\%, outperforming the best baseline, TOPJUDGE, which has an accuracy of 67.31\%.

The superior performance of JurisCTC can be attributed to its ability to effectively leverage the transfer learning from one legal domain to another, capturing the intricate relationships and dependencies within the legal texts.
This is evident from the high macro-recall and macro-F1 scores, indicating that JurisCTC is not only accurate but also consistent in its predictions across different classes.

The high MP scores for GPT-4o, DeepSeek-V3-Chat, and Gemini-1.5-Flash in Criminal Law tasks indicate that these models are very effective at correctly identifying positive instances when they make a positive prediction.
The high MP suggests that these models are conservative in their positive predictions, prioritizing precision over recall.
However, their performance in Civil Law tasks is less consistent, indicating that these models may be more specialized or better tuned for Criminal Law tasks.

While JurisCTC demonstrates superior performance in both Criminal and Civil Law tasks, there are notable areas for improvement.
One key limitation is its precision compared to models like GPT-4o in Criminal Law tasks, where JurisCTC achieves a precision of 75.92\% versus GPT-4o's 83.00\%.
This suggests that JurisCTC, despite its high recall, may produce more false positives, which could be problematic in scenarios requiring high precision.
The reason for this may be that models such as GPT-4o obtain more knowledge from positive civil law data, but the accuracy of these models is not high, which shows that the model proposed in this paper has a balance between learning positive data and negative data.

In general, the results validate the effectiveness of the proposed model in handling complex legal judgment prediction tasks, showcasing its potential for practical applications in the legal domain.

\subsubsection{Ablation Study}

In this section, we conduct an ablation study to evaluate the performance of various models when transferring between Civil Law and Criminal Law tasks.
We compare the baseline BERT model with its variants, including BERT with Unsupervised Data Augmentation (BERT-UDA), BERT with Contrastive Learning (BERT-CL), and JurisCTC.
The results are presented in Tables \ref{tab:ablation_results_civil_to_criminal} and \ref{tab:ablation_results_criminal_to_civil}.

Table \ref{tab:ablation_results_civil_to_criminal} shows the performance of the models trained on Civil Law data and tested on Criminal Law tasks.
The baseline BERT model achieves an accuracy of 43.51\%, highlighting challenges in cross-domain transfer due to significant differences between civil and criminal legal language.
BERT-UDA improves performance to 65.64\%, indicating that data augmentation helps in adapting to the new domain.
BERT-CL achieves an accuracy of 60.61\%, showing that contrastive learning provides some benefit for cross-domain adaptation, though less effective than UDA.
JurisCTC outperforms all models with an accuracy of 76.59\% and shows superior precision, recall, and F1-score, demonstrating its robustness in cross-domain adaptation.

\begin{table}[!t]
\caption{Performance ablation on Civil Law to Criminal Law tasks}
\begin{center}
\begin{tabular}{|l|c|c|c|c|}
\hline
\textbf{Model} & \textbf{Acc.} & \textbf{MP} & \textbf{MR} & \textbf{F1} \\
\hline
BERT & 43.51 & 50.00 & 6.30 & 11.14 \\
BERT-UDA & 65.64 & 71.26 & 65.64 & 68.33 \\
BERT-CL & 60.61 & 65.47 & 63.58 & 65.03 \\
\hline
\textbf{JurisCTC} & \textbf{76.59} & \textbf{75.92} & \textbf{85.75} & \textbf{80.54} \\
\hline
\end{tabular}
\label{tab:ablation_results_civil_to_criminal}
\end{center}
\end{table}

Table \ref{tab:ablation_results_criminal_to_civil} illustrates the performance of models trained on Criminal Law data and tested on Civil Law tasks.
The baseline BERT model achieves an accuracy of 64.37\%, which is notably better compared to the reverse transfer, suggesting that criminal law features might generalize better to civil law contexts.
In this case, BERT-UDA shows a decrease in performance with an accuracy of 57.74\%, indicating the context dependency of UDA's effectiveness.
BERT-CL improves results with an accuracy of 65.87\%, suggesting that contrastive learning may be more beneficial when transferring to civil law. JurisCTC again leads with an accuracy of 78.83\%, along with superior precision, recall, and F1-score, highlighting its superior adaptability across legal domains.

\begin{table}[!t]
\caption{Performance ablation on Criminal Law to Civil Law tasks}
\begin{center}
\begin{tabular}{|l|c|c|c|c|}
\hline
\textbf{Model} & \textbf{Acc.} & \textbf{MP} & \textbf{MR} & \textbf{F1} \\
\hline
BERT & 64.37 & 63.77 & 86.97 & 73.58 \\
BERT-UDA & 57.74 & 56.10 & 55.59 & 55.84 \\
BERT-CL & 65.87 & 66.43 & 85.48 & 74.37 \\
\hline
\textbf{JurisCTC} & \textbf{78.83} & \textbf{76.59} & \textbf{90.61} & \textbf{83.01} \\
\hline
\end{tabular}
\label{tab:ablation_results_criminal_to_civil}
\end{center}
\end{table}

In general, the results of this ablation study demonstrate the effectiveness of our proposed domain-adaptive model, JurisCTC, in the domain of legal judgment prediction (LJP). The model consistently outperforms other variants, showcasing its robustness and adaptability in different legal domains.
This validates the design choices and domain adaptation strategies employed in JurisCTC, highlighting its potential to address the unique challenges of cross-domain transfer in legal tasks.
These findings underscore the model's capability to enhance prediction accuracy and reliability, confirming its suitability for practical applications in the LJP field.

\section{Conclusion}    
We propose JurisCTC, which is a knowledge transfer model for dealing with annotated texts in the legal field and can achieve the transfer of knowledge from different departmental laws.
In the context of a significant decrease in the number of publicly available judgments and the lack of large-scale annotated legal datasets in the Chinese field, we have demonstrated that UDA can learn the logic of a legal application from civil law and apply it to new criminal law in LJP tasks, and significantly improve the prediction accuracy of the target domain.
At the same time, to test the generalizability of the model, we also experimented with learning the logic of legal application from the field of criminal law and applied it to the field of civil law, and the performance of the model was significantly improved.
In short, compared to traditional models, JurisCTC effectively solves the challenges of lengthy and complex legal texts, significantly improves the predictive accuracy of LJP tasks, and enhances the generalization ability of the model.

Future research will investigate the particular attributes of legal language that contribute to the effectiveness of JurisCTC.
We plan to investigate the linguistic features that are most influential in model performance and explore how different domain adaptation strategies enhance our model ability.
This analysis will not only refine our understanding of domain-specific adaptation but also improve the predictive capabilities of AI systems in legal contexts.

\setcounter{page}{5}
\bibliographystyle{IEEEbib}
\bibliography{string}

\end{document}